\documentclass[10pt,twocolumn,letterpaper]{article}

\usepackage{cvpr}
\usepackage{times}
\usepackage{epsfig}
\usepackage{graphicx}
\usepackage{amsmath}
\usepackage{amssymb}
\usepackage{amsthm}
\usepackage{subcaption}

\usepackage[pagebackref=true,breaklinks=true,letterpaper=true,colorlinks,bookmarks=false]{hyperref}
  \providecommand{\corollaryname}{Corollary}
  \providecommand{\definitionname}{Definition}
  \providecommand{\lemmaname}{Lemma}
\providecommand{\theoremname}{Theorem}

\newtheorem{thm}{\protect\theoremname}
  \newtheorem{defn}{\protect\definitionname}
  \newtheorem{lem}{\protect\lemmaname}
\newtheorem{cor}{\protect\corollaryname}
 \cvprfinalcopy % *** Uncomment this line for the final submission

\def\cvprPaperID{1598} % *** Enter the CVPR Paper ID here

\ifcvprfinal\fi
\begin{document}

%%%%%%%%% TITLE
\title{Beyond Gaussian Pyramid: Multi-skip Feature Stacking for Action Recognition}

\author{Zhenzhong Lan, Ming Lin, Xuanchong Li, Alexander G. Hauptmann, Bhiksha Raj \\
School of Computer Science, Carnegie Mellon University\\
{\tt\small lanzhzh, minglin, xcli, alex, bhiksha@cs.cmu.edu}
}

\maketitle
%\thispagestyle{empty}

%%%%%%%%% ABSTRACT
\begin{abstract}
Most state-of-the-art action feature extractors involve differential operators, which act as highpass filters and tend to attenuate low frequency action information. This attenuation introduces bias to the resulting features and generates ill-conditioned feature matrices. The Gaussian Pyramid has been used as a feature enhancing technique that encodes scale-invariant characteristics into the feature space in an attempt to deal with this attenuation. However, at the core of the Gaussian Pyramid is a convolutional smoothing operation, which makes it incapable of generating new features at coarse scales. In order to address this problem, we propose a novel feature enhancing technique called Multi-skIp Feature Stacking (MIFS), which stacks features extracted using a family of differential filters  parameterized with multiple time skips and encodes shift-invariance into the frequency space. MIFS compensates for information lost from using differential operators by recapturing information at coarse scales. This recaptured information allows us to match actions at different speeds and ranges of motion.  We prove that MIFS enhances the learnability of differential-based features exponentially. The resulting feature matrices from MIFS have much smaller conditional numbers and variances than those from conventional methods. Experimental results show significantly improved performance on challenging action recognition and event detection tasks. Specifically, our method exceeds the state-of-the-arts on Hollywood2, UCF101 and UCF50 datasets and is comparable to state-of-the-arts on HMDB51 and Olympics Sports datasets. MIFS can also be used as a speedup strategy for feature extraction with minimal or no accuracy cost. 
\end{abstract}

%%%%%%%%% BODY TEXT

\section{Introduction}

We consider the problem of enhancing video representations for action recognition, which becomes increasingly important for both analyzing human activity itself and as a component for more complex event analysis. As pointed out by Marr \cite{marr1982vision} and Lindeberg \cite{lindeberg1994linear}, visual representations, or visual features, are of utmost importance for a vision system, not only because they are the chief reasons for tremendous progress in the field, but also because they lead to a deeper understanding of the information processing in the human vision system. In fact, most of the qualitative improvements to visual analysis can be attributed to the introduction of improved representations, from SIFT \cite{lowe2004distinctive} to  Deep Convolutional Neural Networks \cite{sermanet2013overfeat}, STIP \cite{laptev2005space} to Dense Trajectory \cite{wang2011action} . A common characteristic of these several generations of visual features is that they all, in some way, benefit from the idea of multi-scale representation, which is generally viewed as an indiscriminately applicable tool that reliably yields an improvement in performance when applied to almost all feature extractors.

\begin{figure}
        \centering
        \begin{subfigure}[b]{0.23\textwidth}
                \includegraphics[width=\textwidth]{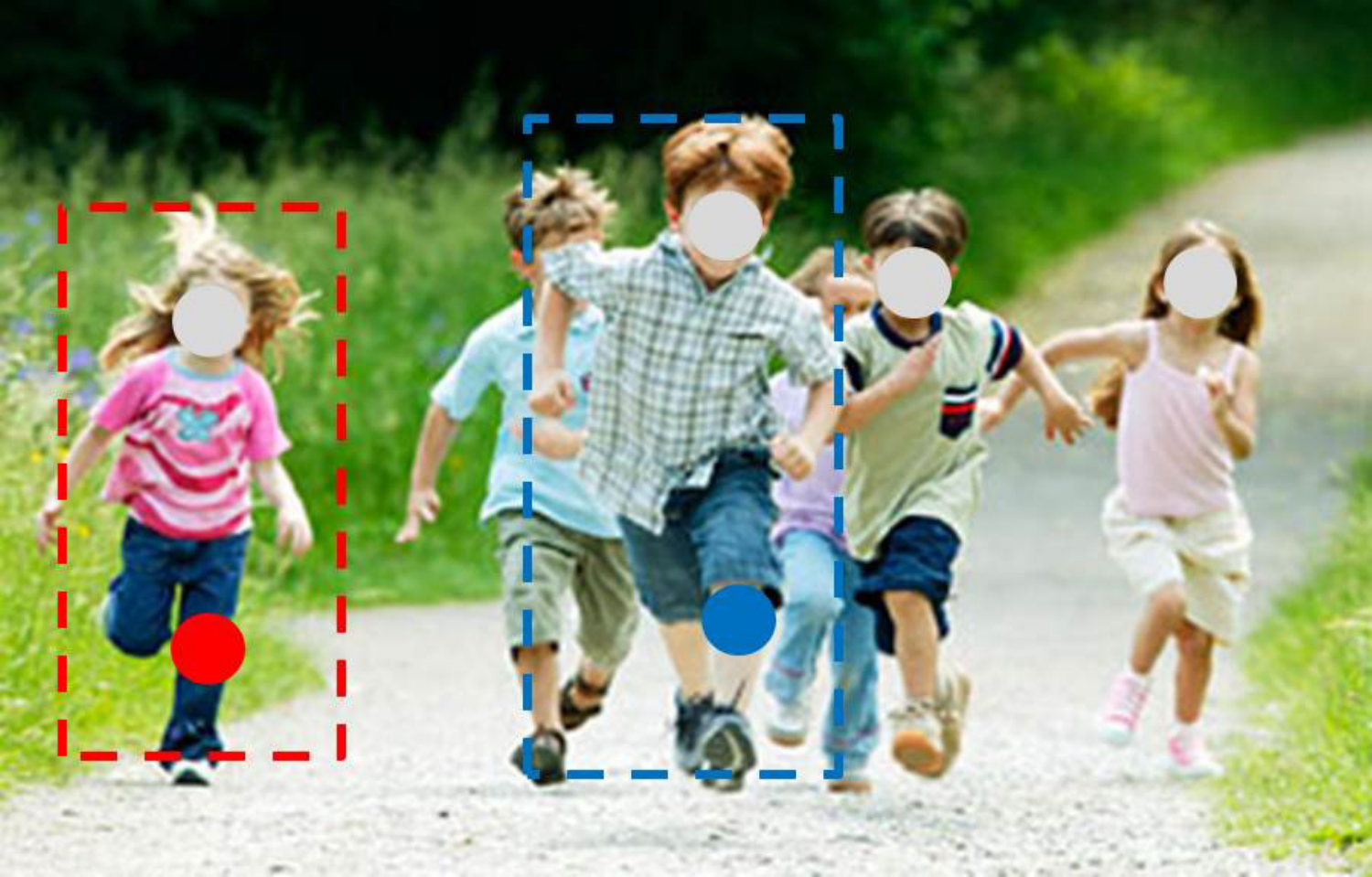}
                \caption{Kids Running}
                \label{fig:kr}
        \end{subfigure}%
        ~ %add desired spacing between images, e. g. ~, \quad, \qquad, \hfill etc.
        \begin{subfigure}[b]{0.23\textwidth}
                \includegraphics[width=\textwidth]{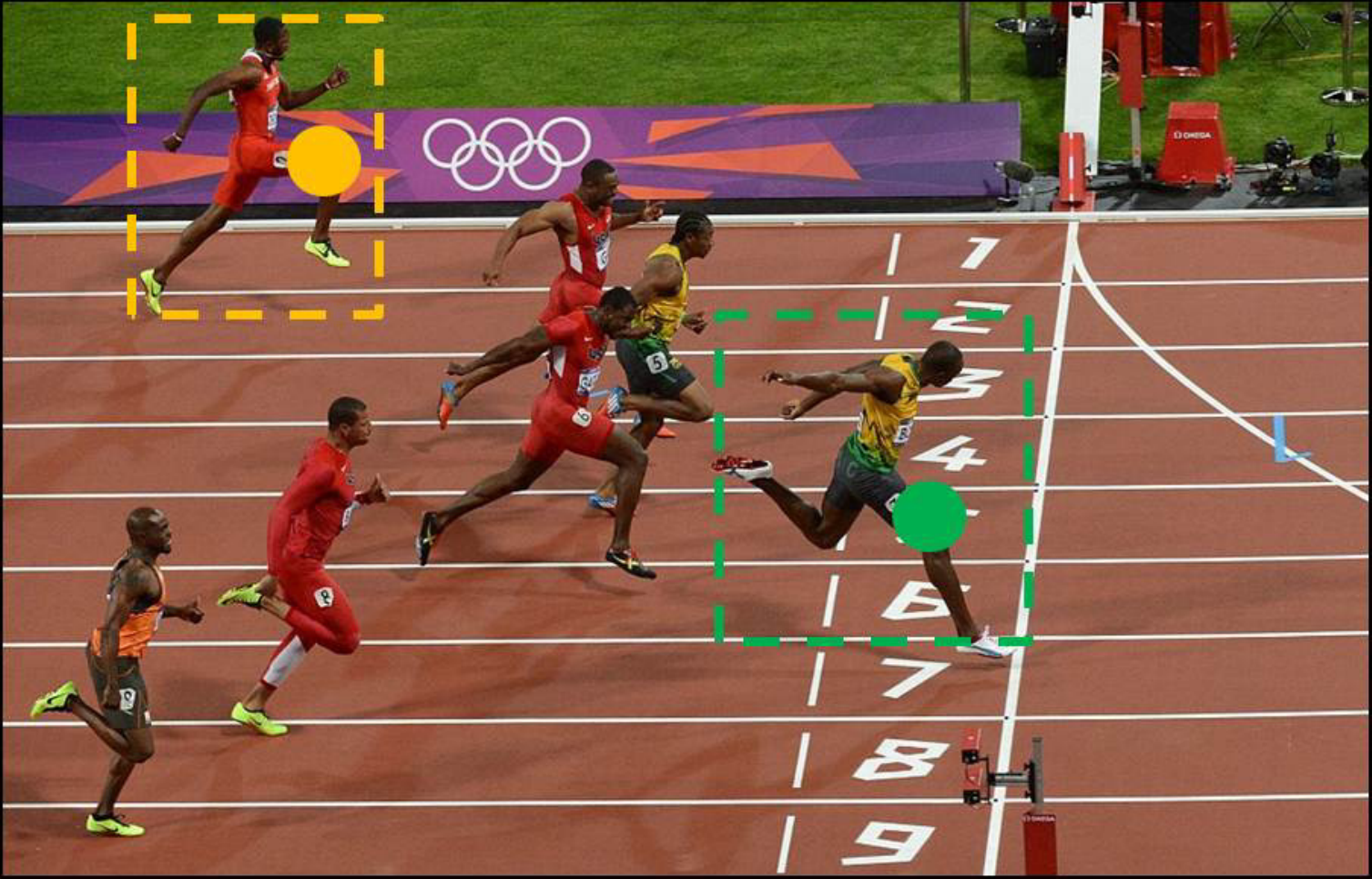}
                \caption{Olympics Running}
                \label{fig:or}
        \end{subfigure}
        ~ %add desired spacing between images, e. g. ~, \quad, \qquad, \hfill etc.
        \begin{subfigure}[b]{0.22\textwidth}
                \includegraphics[width=\textwidth]{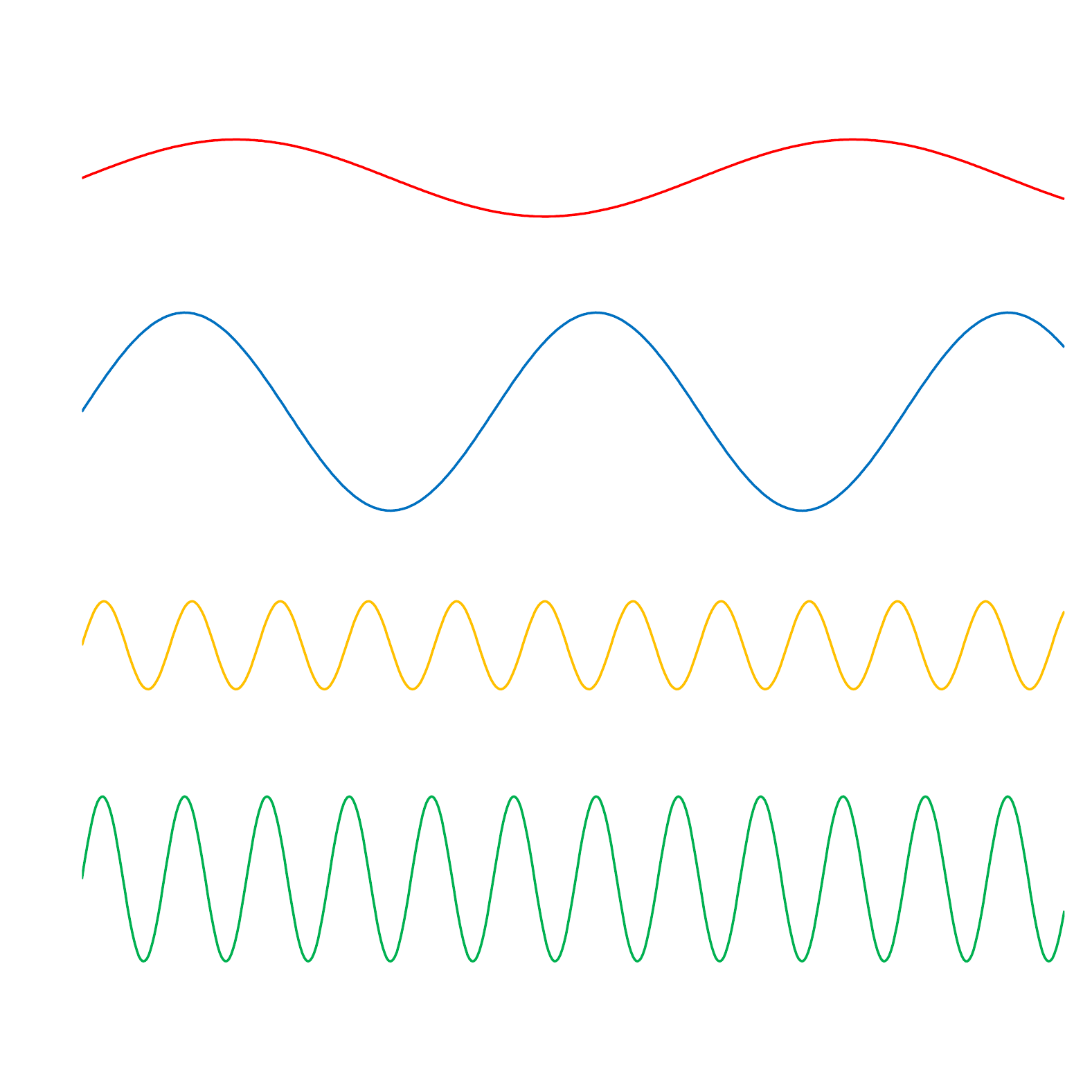}
                \caption{Action signal $S$}
                \label{fig:signal}
        \end{subfigure}
       ~ \quad
        \begin{subfigure}[b]{0.22\textwidth}
                \includegraphics[width=\textwidth]{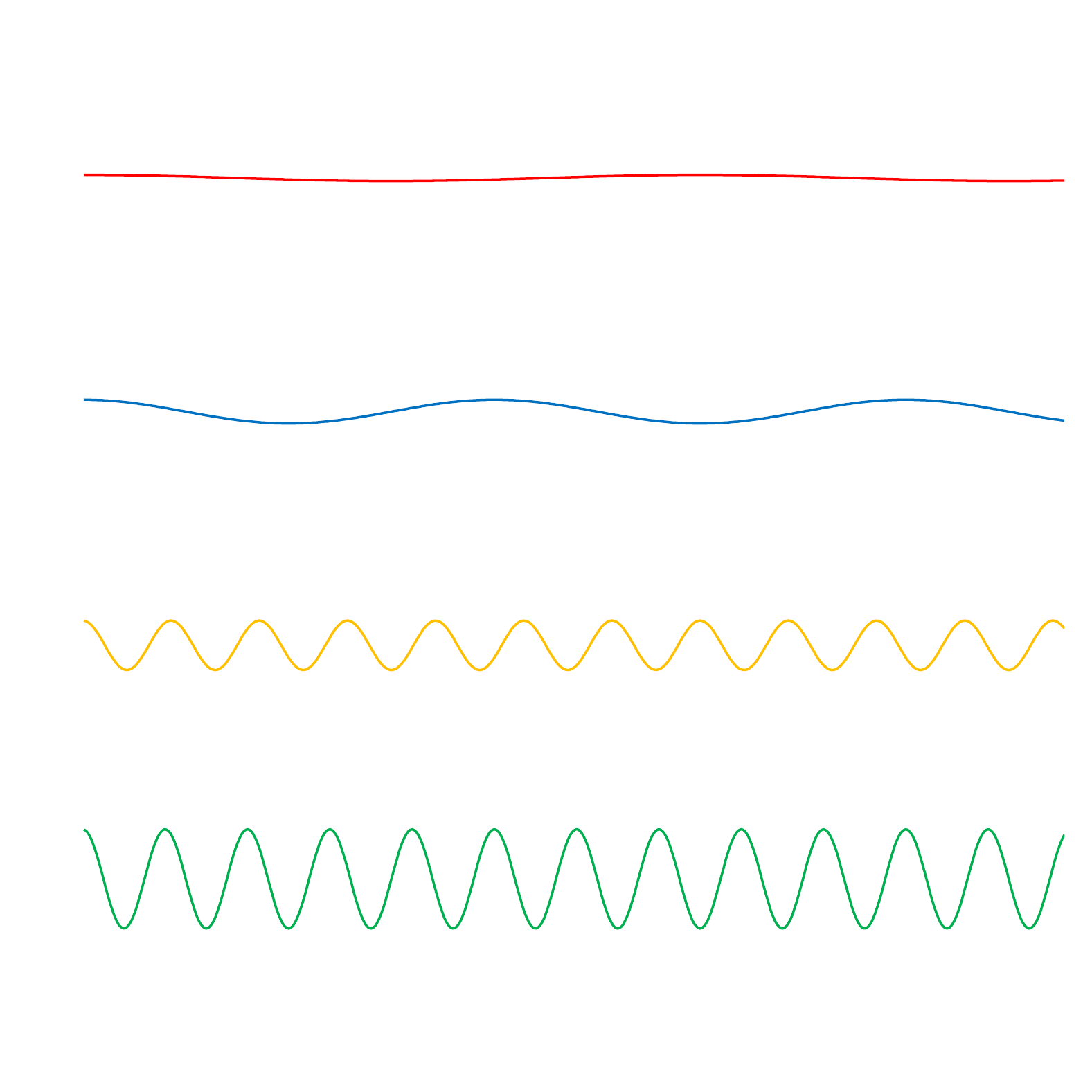}
                \caption{  $\Delta S$ }
                \label{fig:dsignal}
        \end{subfigure}
        \caption{Simplified action signals (\ref{fig:signal}) from "running" actions (\ref{fig:kr},\ref{fig:or}) show dramatically difference among subjects and scenes. With such dramatic differences among action signals, a differential operator with single scale is incapable of covering a full range of action frequency and tend to lose low frequency information (the {\color[rgb]{1,0,0} red} and {\color[rgb]{0,0.44,0.75} cyan} signals) .  }\label{fig:motivation}
\end{figure}

At the core of image multi-scale representation is the requirement that no new detail information should be artificially found at the coarse scale of resolution \cite{koenderink1984structure}. Gaussian Pyramid, a unique solution based on this constraint, generates a family of images where fine-scale information is successively suppressed by Gaussian smoothing.  However, in action recognition, we often desire the opposite requirements.  For example, in generating action features using differential filters, we need coarse-scale  features to:  1) recover the information that has been filtered out by highpass filters at fine scales, e.g., the {\color[rgb]{1,0,0} red} and {\color[rgb]{0,0.44,0.75} cyan} signals in Figure \ref{fig:signal} are likely to be filtered out; 2) generate features at higher frequency for matching similar actions at different speeds and ranges of motion, e.g., the {\color[rgb]{1,0.75,0} orange} and {\color[rgb]{0,0.69,0.31} green} signals in Figure \ref{fig:signal}. Both of these requirements cannot be satisfied with a Gaussian Pyramid representation.

In this work, we introduce a Multi-skIp Feature Stacking (MIFS) representation that works by stacking features extracted by a family of differential filters parameterized with multiple time skips (scales). Our algorithm relies on the idea that by gradually reducing the frame rate, feature extractors with differential filters can extract information about more subtle movements of actions. MIFS has several attractive properties:
\begin{itemize}
\item It is an indiscriminately applicable tool that can be reliably and easily adopted by any feature extractors with differential filters, like Gaussian Pyramid, 
\item It generates features that are shift-invariance in frequency space, hence easier to match similar actions at different speeds and ranges of motion. 
\item It stacks features at multiple frequencies and tends to cover a longer range of action signals， compared to conventional action representations,
\item It generates feature matrices that have smaller conditional numbers and variances hence stronger learnability compared to the conventional original-scale representation based on our theoretical analysis. 
\item It significantly improves the performance of state-of-the-art methods based on experimental results on several real-world benchmark datasets.
\item It exponentially enhances the learnability of the resulting feature matrices. Therefore the required additional number of scales is logarithmic to the bandwidth of the action signals. Empirical studies show that one or two additional scales are enough to recover the information lost  by differential operators. Hence the additional computational cost of MIFS is small.
\item It can be used to as a feature extraction speedup strategy with minimal or no accuracy cost. As shown in our experiments, combining features extracted from videos at lower frame rates (with different time skips) performs better than features from videos at the original frame rate at the same time requires less time to process. 
\end{itemize}

In the remainder of this paper, we start by providing more background information about action recognition and multi-scale presentations. We then describe MIFS in detail, followed by theoretically proving that MIFS improves the learnability of video representations exponentially. After that, an evaluation of our method is performed. Further discussions including potential improvements are given at the end.

\section{Related Work}
There is an extensive body of literature about action recognition; here we just mention a few relevant ones involved with state-of-the-art feature extractors and feature encoding methods. See \cite{aggarwal2011human} for an in-depth survey. In conventional video representations, features and encoding methods are the two chief reasons for considerable progress in the field. Among them, the trajectory based approaches \cite{matikainen2009trajectons,sun2009hierarchical,wang2011action,wang2013action,jiang2012trajectory}, especially the Dense Trajectory method proposed by Wang et al. \cite{wang2011action,wang2013action}, together with the Fisher Vector encoding  \cite{perronnin2010improving}  yields the current state-of-the-art performances on several benchmark action recognition datasets.  Peng et al. \cite{peng2014bag,peng2014action} further improved the performance of Dense Trajectory by increasing the codebook sizes, fusing multiple coding methods and adding a stacked Fisher Vector. Some success has been reported recently using deep convolutional neural networks for action recognition in videos. Karpathy et al. \cite{karpathy2014large} trained a deep convolutional neural network using 1 million weakly labeled YouTube videos and reported a moderate success on using it as a feature extractor. Simonyan $\&$ Zisserman \cite{simonyan2014two} reported a result that is competitive to Improved Dense Trajectory \cite{wang2013action} by training deep convolutional neural networks using both sampled frames and optical flows. MIFS is an indiscriminately applicable tool that can be adopted by all of above mentioned feature extractors. 

Multi-scale representation \cite{adelson1984pyramid, lindeberg1994linear} has been very popular for most image processing tasks such as image compression, image enhancement and object recognition. A multi-scale key-point detector proposed by Lindeberg \cite{lindeberg1993detecting} and used in by Lowe \cite{lowe2004distinctive} to detect scale invariant key points using Laplacian pyramid methods, in which Gaussian smoothing is used iteratively for each pyramid level. Simonyan $\&$ Zisserman \cite{simonyan2014very} reported a significant performance improvement on Imagenet Challenge 2014 by using a multi-scale deep convolutional neural network. In video processing, Space Time Interest Points (STIP) \cite{laptev2005space} extends SIFT to the temporal domain by finding the scale invariant feature points in 3D space. Shao et al. \cite{zhen2013spatio} also try to achieve scale invariance for action recognition using 3-D Laplacian pyramids and 3D Gabor filters. However, without awareness of the fundamental differences between image and video processing, \cite{zhen2013spatio} was not very successful when compared to the state-of-the-art methods.

For lab datasets where human poses or action templates can be reliably estimated, Dynamic Time Warping (DTP) \cite{ darrell1993space}, Hidden Markov Models (HMMs) \cite{yamato1992recognizing} and Dynamic Bayesian Networks (DBNs) \cite{park2004hierarchical} are well studied methods for aligning actions that have speed variation. However, for noisy real-world actions, these methods have not shown themselves to be very robust.

\section{ Multi-skIp Feature Stacking (MIFS)}

\begin{figure*}
\centering
\begin{tabular}{cc}
\includegraphics[width=0.22\textwidth]{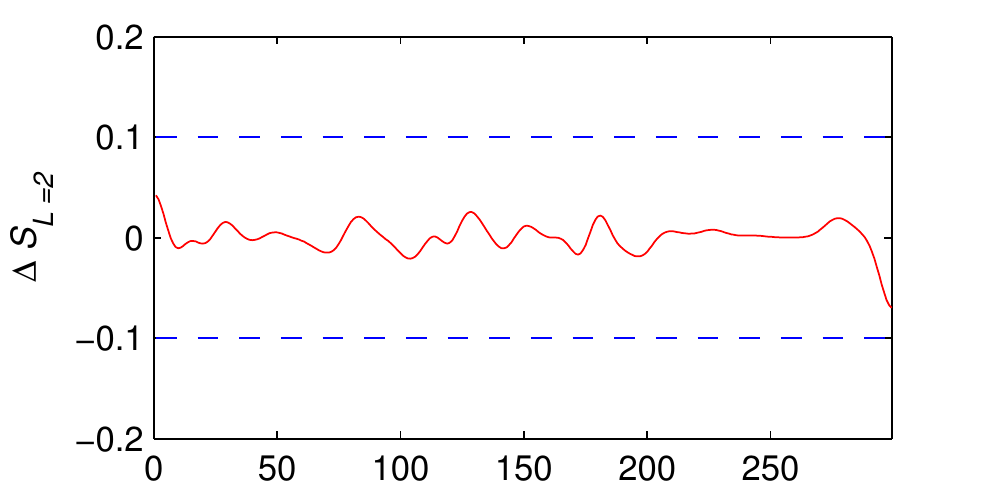} &
\includegraphics[width=0.22\textwidth]{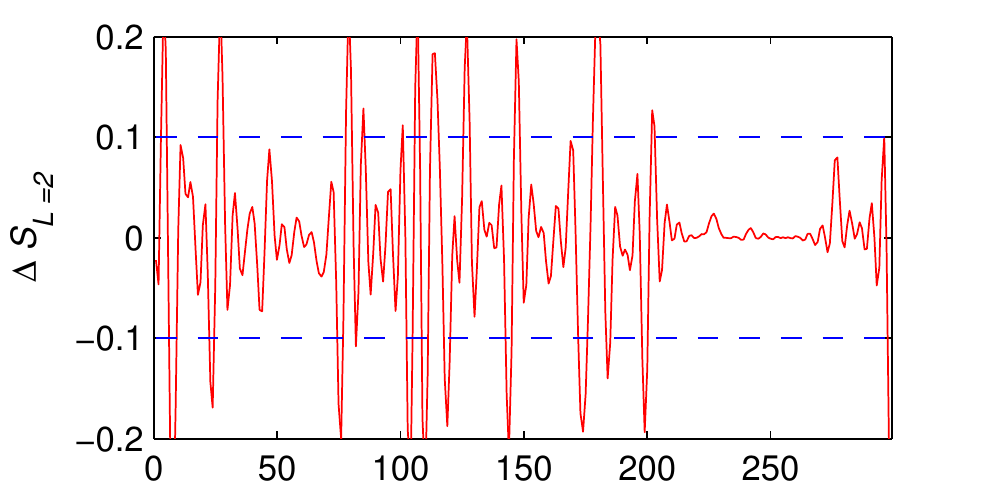}\\
\includegraphics[width=0.44\textwidth]{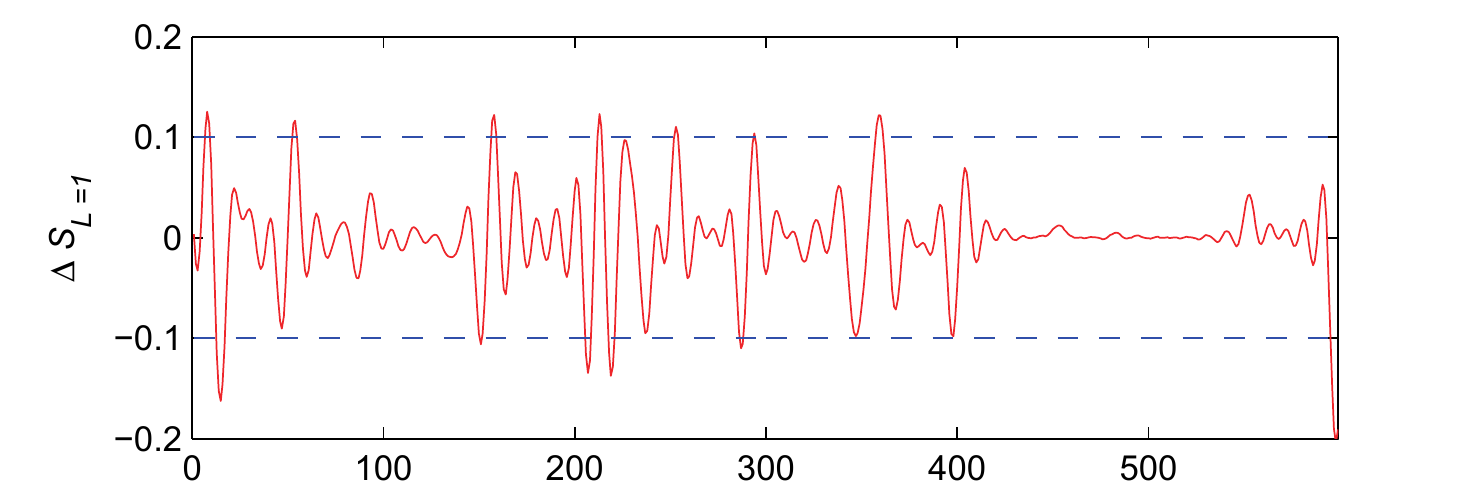} &  
\includegraphics[width=0.44\textwidth]{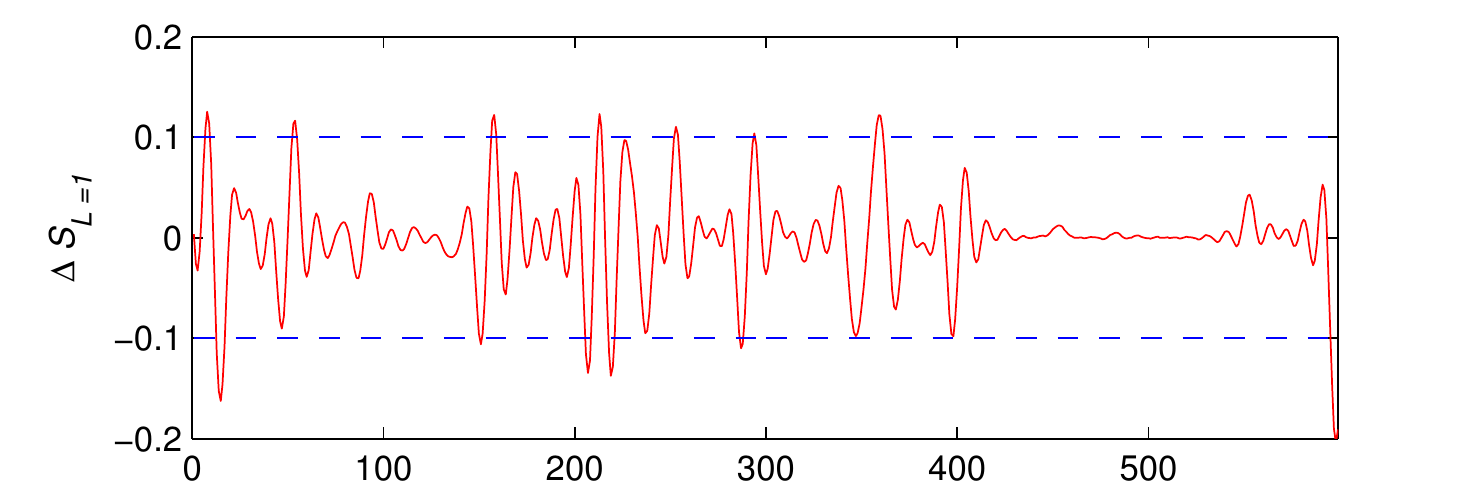}\\
(a) Gaussian Pyramid &(b) MIFS
\end{tabular}
\caption{Comparison of Gaussian Pyramid and MIFS for a real action signal. The left figure (a) shows that as the level ($L$) goes higher (from 1 to 2), the resulting features ($\Delta S$) from a differential operator become less prominent. So once a feature has been filtered out (assume the threshold for a feature to be represented is 0.1), it cannot be recovered by higher level features under the Gaussian Pyramid framework. The right figure (b) shows that under MIFS, the features ($\Delta S$) become more prominent as the levels go higher and can represent those signals that have been filtered out at low levels.}\label{fig:msr}
\end{figure*}

We now formalize our notation. For the present discussion a video $X$ is just a real function of three variables:
\begin{align}
X = X(x, y, t).
\end{align}
The normalized coordinates $(x, y, t) \in R^3$ are the  Cartesian coordinates
of the video space.
Since we focus on the temporal domain, we omit $(x,y)$ in further discussion and
denote a video as $X(t)$. The length of the video is assumed to be normalized,
that is $t\in[0,1]$.
In our model, the content of a video is generated by a linear mixture of  $k$ latent signals:
\begin{align} \label{eq:video-latent-signal-model}
  \bar{X}=[\bar{\mathbf{x}}_{1},\bar{\mathbf{x}}_{2},\cdots,\bar{\mathbf{x}}_{k}] \ .
\end{align}

The mixing weight of each latent action signal
$\bar{\mathbf{x}}_{i}$ at time $t$ is denoted as $\alpha_{i}(t)$. Therefore, a
given video is generated as
\begin{align}
  X(t)=&\bar{X}\boldsymbol{\alpha}(t)+\boldsymbol{\epsilon}(t) \\
  \boldsymbol{\alpha}(t)=&[\alpha_{1}(t),\alpha_{2}(t),\cdots,\alpha_{k}(t)]{}^{\mathrm{T}}~.
\end{align}
where $\boldsymbol{\epsilon}(t)$ is additive subgaussian noise with noise level
$\sigma$. We assume $\forall i $,
\begin{align}
|\alpha_{i}(t)|\leq 1 & \quad  \mathrm{E}_{t}\{\alpha_{i}(t)\}=0 \\
\mathrm{E}_{t}\{\alpha_{i}(t)^{2}\}\leq1 &\quad  \mathrm{E}_{t}\{ \alpha_{i}(t) \times \alpha_{j}(t)|_{i\neq j}\} = 0   ~.
\end{align}

The feature extractor is assumed to be modeled as a differential operator
$\mathcal{F}[\cdot,\tau]$ parameterized with time skip $\tau$. Given a fixed
$\tau$, the feature extractor $\mathcal{F}[X(t),\tau]$ generates $T
= \left\lfloor 1/\tau\right\rfloor $ features.
\[
\mathcal{F}[X(t),\tau]=[\mathbf{f}(t_{1},\tau),\mathbf{f}(t_{2},\tau),\cdots,\mathbf{f}(t_{T},\tau)]~.
\]
where $t_{1},t_{2},\cdots,t_{T}$ are uniformly sampled on $[0,1]$.
The $i$-th feature vector $\mathbf{f}(t_{i},\tau)$  is generated by
\begin{align}
f(t_{i},\tau) & = X(t_i+\tau)-X(t_i) \\
& =  \bar{X}\times(\boldsymbol{\alpha}(t_i+\tau)-\boldsymbol{\alpha}(t_i))+\boldsymbol{\epsilon}(t_i+\tau)-\boldsymbol{\epsilon}(t_i)~. \nonumber
\end{align}

We can rewrite the feature matrix  as
\[
\mathcal{F}[X(t),\tau]=\bar{X}P+\sum_{i=1}^{T}\boldsymbol{\epsilon}(t_{i}+\tau)-\boldsymbol{\epsilon}(t_{i})
\]
 where $P$ is a $k\times T$ matrix, $P_{i,j}=\boldsymbol{\alpha}_{i}(t_{j}+\tau)-\boldsymbol{\alpha}_{i}(t_j)$.
 
Most action feature extractors are different versions of $\mathcal{F}$. For
example, STIP \cite{laptev2005space} and Dense Trajectory
\cite{wang2011action} can be derived from $\mathcal{F}[\cdot, \frac{1}{K}]$,
where $K$ is the number of frames in the video.

MIFS stacks multiple $\mathcal{F}[X(t),\tau]$ with different $\tau$. By stacking multiple features with different frequencies, MIFS seeks invariance in the frequency domain via
resampling in the time domain. Figure \ref{fig:msr} shows the difference of Gaussian
Pyramid and MIFS for a real signal from an unconstrained video. It is
clear that, because of smoothing, Gaussian Pyramid fails to recover signals once they have been filtered out. As the levels go higher, the feature generated by Gaussian Pyramids can only become weaker. While in MIFS, the generated features become more prominent and can be recovered as the levels go higher. 

\section{The Learnability of MIFS}

In this section, we first show that under model
Eq.~\eqref{eq:video-latent-signal-model}, the standard feature extraction
method cannot produce a  feature matrix conditioned well enough. Then we show that MIFS improves the condition number of the extracted feature matrix
exponentially. One of the key novelties of the MIFS is that it
also reduces the uncertainty of the feature matrix simultaneously. This reduction is not
possible in a naive approach.

\subsection{Condition Number of $P$ under a Fixed $\tau$}

In this subsection, we will prove, based on the Matrix Bernstein's  Inequality
\cite{tropp2012user},  that
the condition number of $P$ is not necessarily a small number. 

In static feature extractors  such as SIFT, the weight coefficient
matrix $\boldsymbol{\alpha}$ is independent of $t$. While in a video stream, the
action signal is dynamic in $t$. To measure the dynamic of an action signal, we introduce $\gamma_{i}$ as an
index.
\begin{defn}
\label{def:alpha-independent}A latent action signal is $\gamma$
dynamic, if given a non-negative constant $c\in[0,1]$, $\forall\tau\in(0,1]$,
\[
1-(1+c)\exp(-\gamma/\tau)\leq\mathrm{E}_{t}|\alpha(t)\alpha(t+\tau)|\leq1-\exp(-\gamma/\tau)~,
\]
provided $1-(1+c)\exp(-\gamma/\tau) \geq 0. $
\end{defn}
The value $\gamma$ measures how fast the coefficient $\alpha(t)$ varies along
time $t$. Here we take the exponential function by assuming the correlation
between $\alpha(t+\tau)$ and $\alpha(t)$ to be at least subgaussian. If in a
given video, the $i$-th action signal is a high frequency component, then its
coefficient $\boldsymbol{\alpha}_{i}(t)$ will behave like a random number for 
time skip $\tau$. Therefore, we would expect that the correlation between
$\boldsymbol{\alpha}_{i}(t)$ and $\boldsymbol{\alpha}_{i}(t+\tau)$ is close to
0. Or if the action signal is a low frequency component, the correlation indicator $\gamma$ should
be close to $1$. For the sake of simplicity, we rearrange latent action
signal $\bar{X}$ by their frequency to have
$\gamma_{1}\leq\gamma_{2}\leq\cdots\leq\gamma_{k}$.

 In a learning problem, we hope the feature matrix $\mathcal{F}[X(t),\tau]$ to be
 well-conditioned. Given feature matrix $\mathcal{F}[X(t),\tau]$, we can
 recover $\bar{X}$ by various methods, such as subspace clustering. The
 sampling complexity of any recovery algorithm depends on the condition number
 of $P$. Clearly when $P$ is ill--conditioned, we require a large number of
 training examples to estimate $\bar{X}$. The learnability of
 $\mathcal{F}[X(t),\tau]$ depends on its condition number \cite{chu2005foundations} which in return
 depends on $P$ again. In the following, we will prove that for a fixed time
 skip $\tau$, $P$ is not necessarily well conditioned. Therefore the learnability of $\mathcal{F}[X(t),\tau]$ is suboptimal. The intuition
behind our proof is that when an action signal has a large $\gamma$, then a
small time skip $\tau$ will make the coefficient of that signal close to
zero. Therefore, $P$ is ill-conditioned.
Formally, we have the following
theorem to bound the condition number  $\beta(PP{}^{\mathrm{T}})$ of $PP{}^{\mathrm{T}}$(see the proof in supplementary materials). 

\begin{thm}
\label{thm:condition-number-P-under-fixed-tao} Given a fixed time
skip $\tau$, with probability at least $1-\delta$, the condition number
$\beta(PP{}^{\mathrm{T}})$ is bounded by
\begin{align}
  &\beta(PP{}^{\mathrm{T}})\leq\frac{(1+c)\exp(-\gamma_{1}/\tau)+\Delta_{\tau}}{\exp(-\gamma_{k}/\tau)-\Delta_{\tau}} \\
  &\beta(PP{}^{\mathrm{T}})\geq \frac{(1+c)\exp(-\gamma_{1}/\tau)-\Delta_{\tau}}{\exp(-\gamma_{k}/\tau)+\Delta_{\tau}} ~.
\end{align}
where
\begin{align}
\Delta_{\tau}=2\sqrt{k\frac{1}{T}(1+c)\log(2k/\delta)}
\end{align}
provided the number of feature points is
\begin{align}
T\geq\frac{1}{9(1+c)}k\log(2k/\delta) ~.  
\end{align}
\end{thm}

Theorem \ref{thm:condition-number-P-under-fixed-tao} shows that when the number of features $T$ is large enough, the condition number $\beta(PP{}^{\mathrm{T}})$ is a random number concentrated around its expectation $(1+c)\frac{\exp(-\gamma_{1}/\tau)}{\exp(-\gamma_{k}/\tau)}$.
Since $\gamma_{1} \ll \gamma_{k}$, the numerator is much greater than the denominator when $\tau$ is fixed. Since our proof is based on Bernstein's Inequality, the upper bound is  tight. This forces $\beta(PP{}^{\mathrm{T}})$ to be a relatively large value. More specifically, the following corollary
shows that when $\gamma_{k}$ is linear to $\gamma_{1}$, $\beta(PP{}^{\mathrm{T}})$
is exponentially large in expectation. 
\begin{cor}
\label{def:cor}
When $\gamma_{k}\geq (M+1)\gamma_{1}$,
\begin{align}
\mathrm{E}\{\beta(PP{}^{\mathrm{T}})\} \geq
(1+c)[\exp(\frac{\gamma_{1}}{\tau})]^{M} \geq (1+c)(1+\frac{\gamma_{1}}{\tau})^{M}~.  
\end{align}
\end{cor}

Corollary \ref{def:cor} shows that when the actions in the video span across a
vast dynamic range (large M), the feature extractor with single $\tau$ tends
to have ill-conditioned feature matrices. A naive solution to this problem is to
increase $\tau$ to reduce the condition number in expection. However, this will
increase the variance $\Delta_\tau$ of $\beta(PP^T)$ because of a smaller number of features. In practice,  a large $\tau$  also increases the  difficulty in optical flow calculation and tracking. Hence, as will also be observed in
our experiments, choosing a good $\tau$ can be fairly difficult. Intuitively speaking, selecting $\tau$ is a trade-off between feature bias and variance.  A feature extractor with a large $\tau$ covers a long range of action signals but with less feature points hence generates features with small bias but large variance. Similarly, a feature extractor with a small $\tau$ will generate features with large bias but small variance.

\subsection{Condition Number of P under Multiple $\tau$}
\label{p3}

From Theorem \ref{thm:condition-number-P-under-fixed-tao}, to make $PP{}^{\mathrm{T}}$ well-conditioned, we need $\tau$ as large as possible. However, when $\tau$ is too large, we cannot sample enough high quality feature points, therefore, the variance in $PP{}^{\mathrm{T}}$ will increase.
To address this problem, we propose to use MIFS, which incrementally enlarges the time skip $\tau$ , then stacks
all features under various $\tau_{i}$ to form a feature matrix. Hopefully, by
increasing $\tau$, we improve the condition number $\beta(PP{}^{\mathrm{T}})$
and by stacking, we sample enough features to reduce the variance.

Assuming we have features extracted from $\{\tau,2\tau,\cdots m\tau\}$. For $i\tau$ skip, the number of extracted features is
$T_{i} = \left\lfloor 1/(i\tau)\right\rfloor $. The following theorem bounds
the condition number of MIFS (see the proof in supplementary materials). 
\begin{thm}
\label{thm:condition-number-MSP} With probability at least $1-\delta$, the condition number of $PP{}^{\mathrm{T}}$
in the MIFS is bounded by
\begin{align}
\beta(PP{}^{\mathrm{T}})\leq  \frac{\sum_{i}\frac{T_{i}}{T}2(1+c)\exp(-\gamma_{1}/\tau_{i})+\Delta_{\tau}}{\sum_{i}\frac{T_{i}}{T}2\exp(-\gamma_{k}/\tau_{i})-\Delta_{\tau}}~.  
\end{align}
where
\begin{align}
\Delta_{\tau}\leq  2\sqrt{k\frac{1}{\sum_iT_i}(1+c)\log(2k/\delta)}~.
\end{align}
\end{thm}

Theorem \ref{thm:condition-number-MSP} shows that, in the MIFS, the expected condition number $\beta(PP{}^{\mathrm{T}})$ is roughly the weighted
average of condition numbers under various $\tau_{i}$. Since
$\tau_{i+1}>\tau_{i}$, the condition number under $\tau_{i+1}$ is smaller than
the one under $\tau_{i}$. Therefore, the condition number is reduced as we
expected. What's nicer is that the variance component $\Delta_{\tau}$ is
actually on order of $1/\sqrt{\sum_{i}T_{i}}$, which is also much smaller than a single $\tau$ scenario. In summary, we prove:
\begin{quote}
  The MIFS representation improves the learnability of differential
  feature extractors because it reduces the expectation and variance of
  condition number $\beta(PP{}^{\mathrm{T}})$ simultaneously.
\end{quote}

\section{Experiments}
We examine our hypothesis and the proposed MIFS representation on two tasks: action recognition and event detection. The experimental results show that MIFS representations outperform conventional original-scale representations on seven real-world challenging datasets. 

Improved Dense Trajectory with Fisher Vector encoding \cite{wang2013action} represents the current state-of-the-arts for most real-world action recognition datasets. Therefore, we use it to evaluate our method. Note that although we use Improved Dense Trajectory, our methods can be applied to any local features that use differential filters, e.g., STIP \cite{laptev2005space}.

\begin{figure*}
        \centering
        \begin{subfigure}[b]{0.3\textwidth}
                \includegraphics[width=\textwidth]{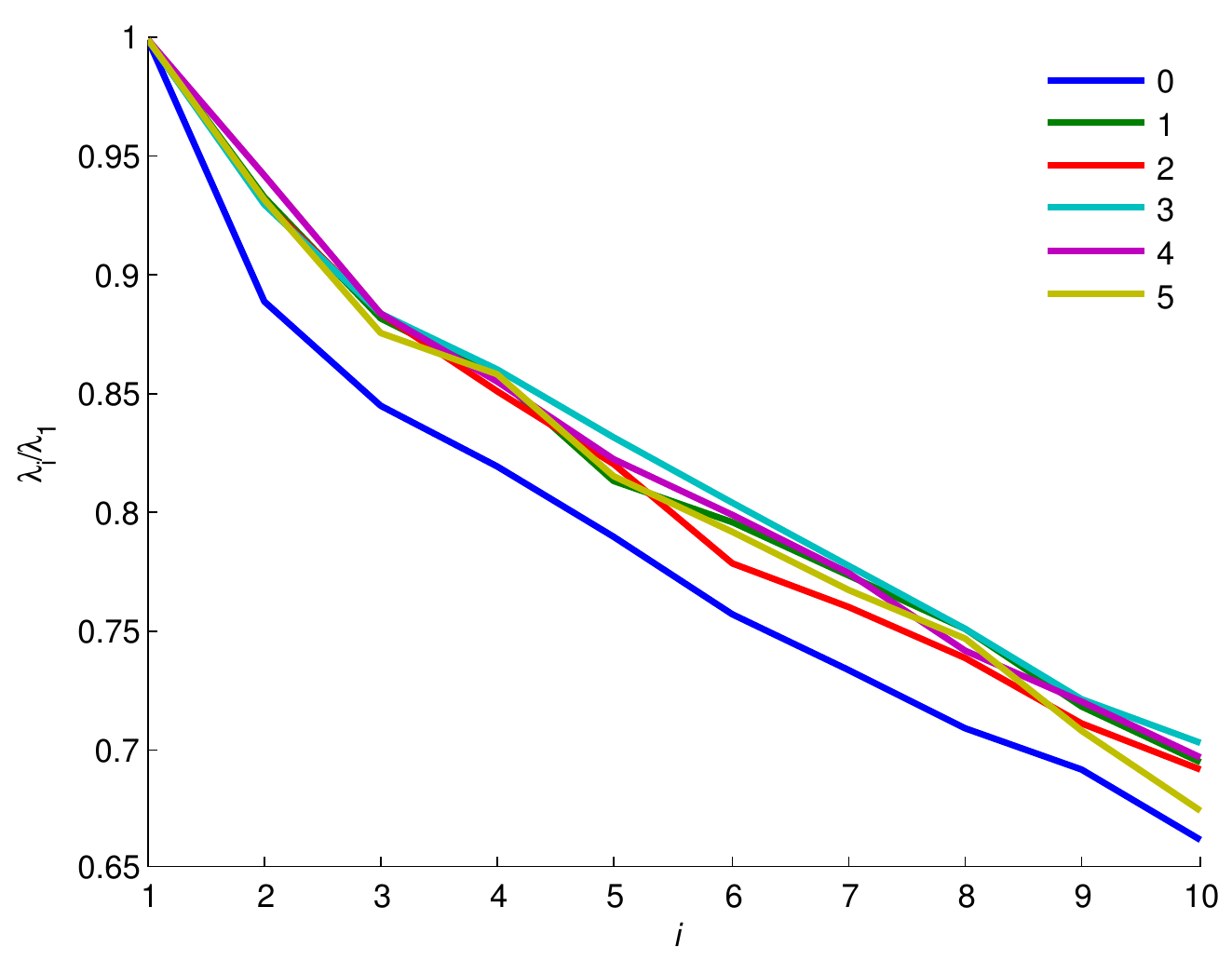}
                \caption{HMDB51}
                \label{fig:lhmdb}
        \end{subfigure}%
~
        \begin{subfigure}[b]{0.3\textwidth}
                \includegraphics[width=\textwidth]{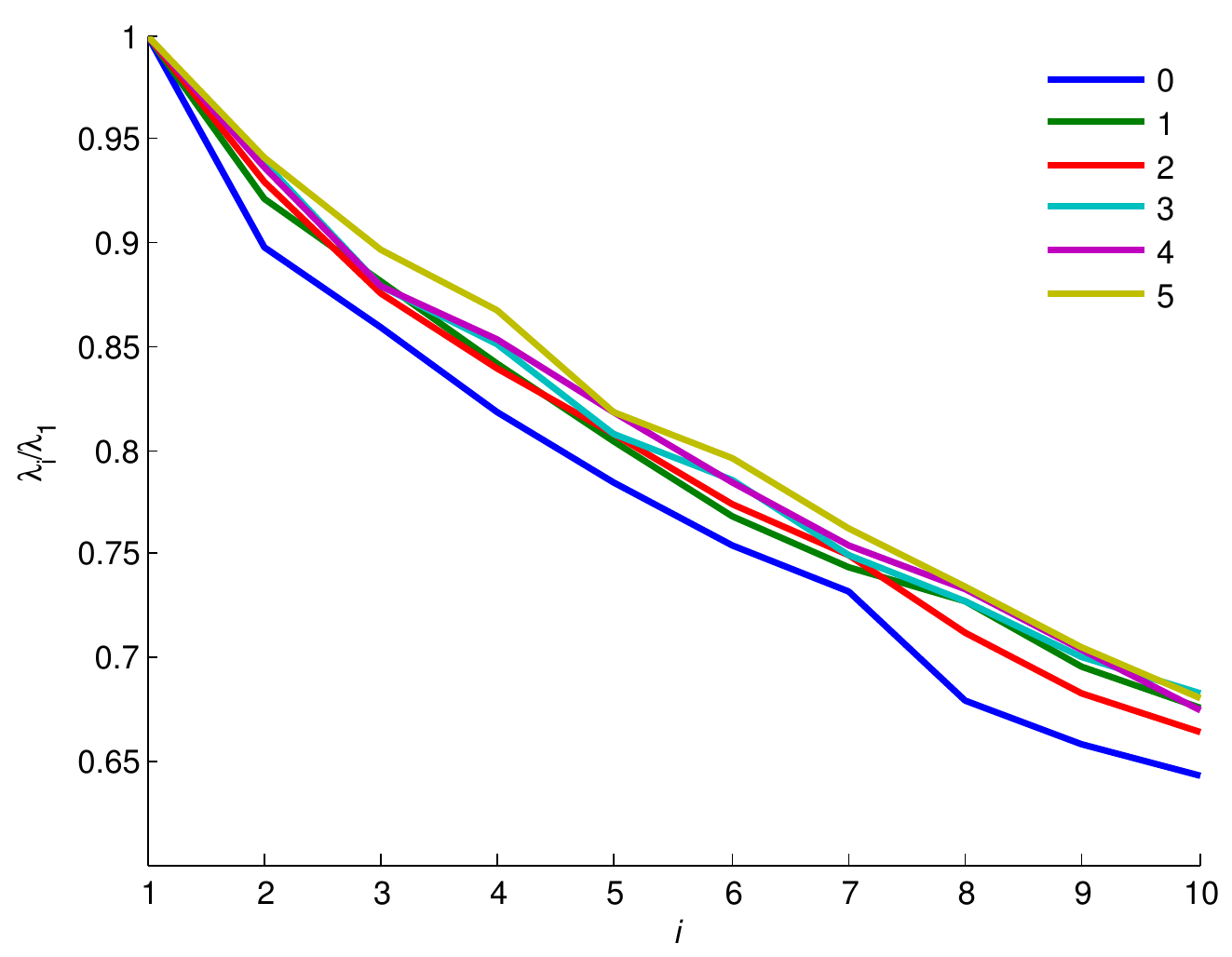}
                \caption{Hollywood2}
                \label{afig:h2}
        \end{subfigure}
~                
        \begin{subfigure}[b]{0.3\textwidth}
                \includegraphics[width=\textwidth]{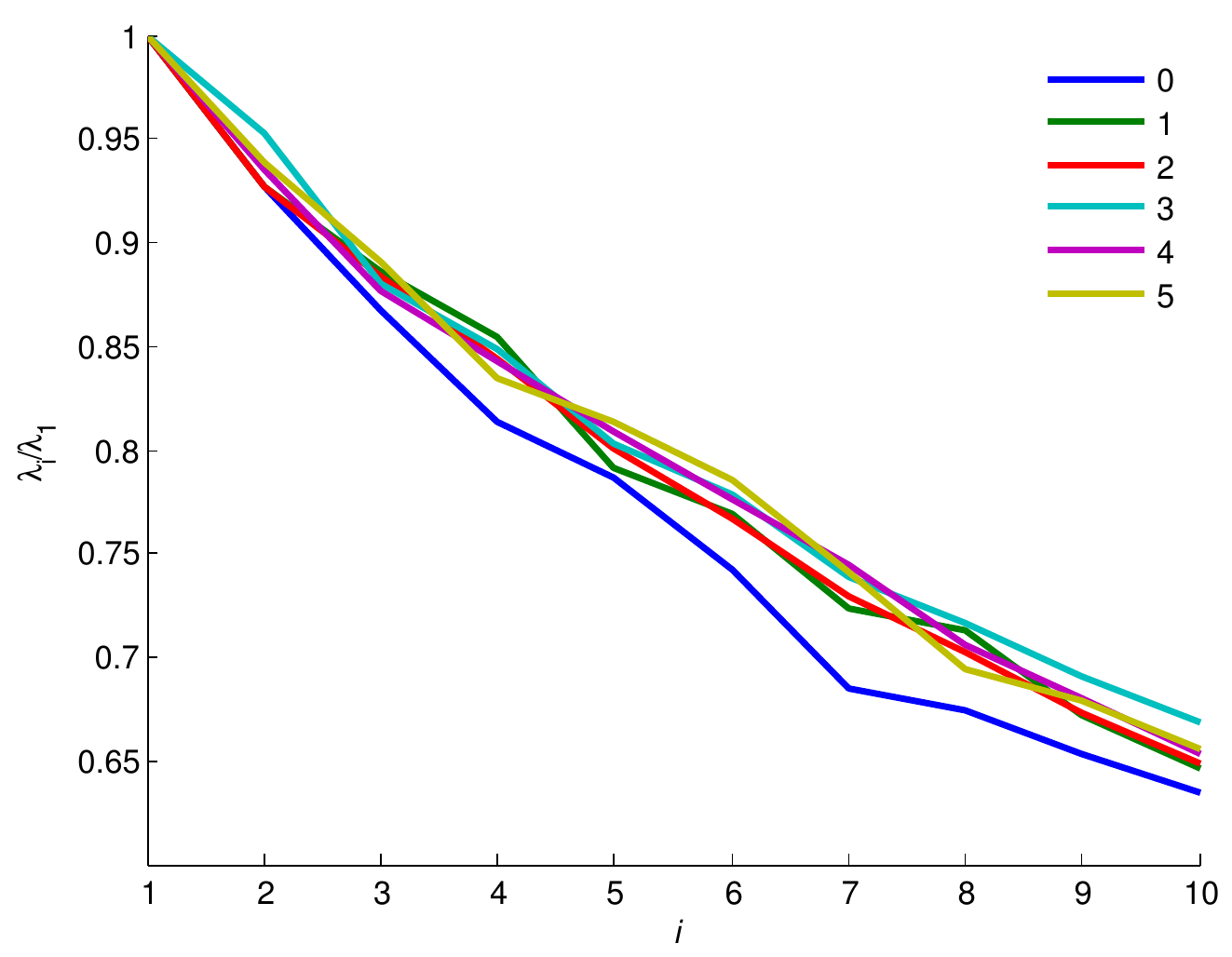}
                \caption{UCF101}
                \label{fig:ucf101}
        \end{subfigure}
        \caption{The decaying trends of singular values of feature matrices for HMDB51,  Hollywood and UCF101 Datasets. 0 to 5 indicate the MIFS level and $i$ indicates the $i$th singular value. From all three datasets, we can see that MIFS representations do have a slower singular value decaying trend compared to conventional representations ({\color[rgb]{0,0,1} blue} lines). }\label{fig:trend}
\end{figure*}

\subsection{Action Recognition}
\label{action_recognition}
\paragraph{Problem Formulation}
The goal of this task is to recognize human actions in short clips of videos. 
\paragraph{Datasets}Five representative datasets are used: The HMDB51 dataset \cite{kuehne2011hmdb} has 51 action classes and 6766 video clips extracted from digitized movies and YouTube. \cite{kuehne2011hmdb} provides both original videos and stabilized ones. We only use original videos in this paper and standard splits with MAcc (mean accuracy) are used to evaluate the performance. The Hollywood2 dataset \cite{marszalek2009actions} contains 12 action classes and 1707 video clips that are collected from 69 different Hollywood movies. We use the standard splits with training and test videos provided by \cite{marszalek2009actions}.   Mean average precision (MAP) is used to evaluate this dataset because multiple labels can be assigned to one video clip. The UCF101 dataset \cite{soomro2012ucf101} has 101 action classes spanning over 13320 YouTube videos clips. We use the standard splits with training and test videos provided by \cite{soomro2012ucf101} and MAcc is reported. The UCF50 dataset \cite{reddy2013recognizing} has 50 action classes spanning over 6618 YouTube videos clips that can be split into 25 groups. The video clips in the same group are generally very similar in background. Leave-one-group-out cross-validation as recommended by \cite{reddy2013recognizing} is used and mean accuracy (mAcc) over all classes and all groups is reported. The Olympic Sports dataset \cite{niebles2010modeling} consists of 16 athletes practicing sports, represented by a total of 783 video clips. We use standard splits with 649 training clips and 134 test clips and report mAP as in \cite{niebles2010modeling} for comparison purposes. 
\paragraph{Experimental Setting}
Improved Dense Trajectory features are extracted using 15 frame tracking, camera motion stabilization and RootSIFT normalization and described by Trajectory, HOG, HOF, MBHx and MBHy descriptors. We use PCA to reduce the dimensionality of these descriptors by a factor of two. After reduction, we augmented the descriptors with three dimensional normalized  location information. The only difference between MIFS and other conventional methods is that instead of using feature points extracted from one time scale, we extract and stack all the raw feature points from different scales together before encoding. For Fisher Vector encoding,  we map the raw descriptors into a  Gaussian Mixture Model with 256 Gaussians trained from a set of randomly sampled 256000 data points. Power and L2 normalization are also used before concatenating different types of descriptors into a video based representation. Another L2 normalization is used after the concatenation. This renormalization brings us about $1\%$ improvement over the baseline method on most of the datasets except Olympics Sports.  For classification, we use a linear SVM classifier with a fixed $C=100$ as recommended by \cite{wang2013action} and the one-versus-all approach is used for multi-class classification scenario. 

\paragraph{Results}\hspace{0pt}

We first examine if it is true that the conditional number $\beta(PP^{\mathrm{T}})$ is improved by MIFS. However, it would not be meaningful to compute $\beta(PP^{\mathrm{T}})$ directly because we have noise
$\boldsymbol{\epsilon}$ in $\mathcal{F}[X(t),\tau_i]$ and the smallest singular value $\lambda_{min}$ is in noise space . A workaround is to examine the decaying speed of singular values of the feature matrix. The
singular values are normalized by dividing the maximum singlular value $\lambda_{\max}$.
We only plot the top $10$ singular values, since the subspace spanned by the
small singular values is noise space. Clearly, when MIFS improves the
learnability, we should get a slower decaying curve of the top $k$ singular
values. Shown in Figure \ref{fig:trend} are the trends of
$\frac{\lambda_i}{\lambda_{max}}$ on the first three datasets: HMDB51, Hollywood2 and UCF101. On all three datasets, the singular values of MIFS decrease slower than the conventional one (0). It is also interesting to see  that by having one or two additional levels, we have already exploited most of the potential improvement.

We further examine how performance changes with respect to the MIFS level, as shown in Table \ref{tab:scalePyramid}. First, let us compare the performance of L=0 to the standard location-insentative feature representation. Our performance on HMDB51, Hollywood2, UCF101 and UCF50 datasets are $62.1\%$ MAcc, $67.0\%$ MAP, $87.3\%$ MAcc and $93.0\%$ respectively. These numbers are higher than Wang \& Schmid \cite{wang2013action}'s results, which are $57.2\%$, $64.3\%$, $85.9\%$ and $91.2\%$, respectively. This improvement is largely because of our location sensitive feature representation and the renormalization. Next, let us check the behavior of MIFS. For completeness, we list both single-scale and stacking performance. For single-scale performance, we observe that for HMDB51, its performance increases from $62.1\%$ to $63.1\%$ and then decrease rapidly, similar patterns can be seen in other datasets except some of them do not increase at $L=1$. These results consist with our observation that different actions need different scale ranges. They also substantiate our proof that selecting time interval $\tau$ is a trade-off between the feature bias and its variance. If computational cost is critical, then we can choose to only extract higher single scale features but suffering minimal or no accuracy lost and enjoying large computational reduction. Now let us compare MIFS with single-scale representation. We observe that for MIFS representations, although there is still a bias and variance trade-off as in single-scale representations for different levels, they all perform better than single-scale representation and the performance decreasing points are later than those in the single-scale representations. We also observe that for MIFS representations, most of the performance improvement comes from $L=1$ and $L=2$, which supports what we observed in Figure \ref{fig:trend} that, in practice, having one or two more scales is enough to recover most of lost information due to the differential operations. Higher scale features become less reliable due to the increasing difficulty in optical flow estimation and tracking.  It is also interesting to observe that HMDB51 enjoys a higher performance improvement from MIFS than the other four datasets have. We believe that the main reason is that HMDB dataset is a mixture of videos from two sources: Youtube and movie, which results in larger action velocity range than pure movie videos or Youtube in other datasets.

\begin{table*}
\centering
\begin{tabular}{|c | c |c |c |c |c| c| c|c |c |c | }
\hline
  & \multicolumn{2}{|c|}{HMDB51  } & \multicolumn{2}{|c|}{ Hollywood2 } &\multicolumn{2}{|c|}{UCF101}   & \multicolumn{2}{|c|}{UCF50  } & \multicolumn{2}{|c|}{ Olympics Sports } \\
    & \multicolumn{2}{|c|}{ (MAcc$\%$) } & \multicolumn{2}{|c|}{(MAP$\%$)} &\multicolumn{2}{|c|}{(MAcc$\%$)} & \multicolumn{2}{|c|}{(MAcc$\%$) } & \multicolumn{2}{|c|}{  (MAP$\%$)} \\    \hline
  L& single-scale & MIFS & single-scale & MIFS & single-scale & MIFS & single-scale & MIFS & single-scale & MIFS\\\hline 

 0 &   62.1 &                   & 67.0 &                & 87.3 & & 93.0 &                   & 89.8 & \\
 1 &   63.1 & 63.8              & 66.4 & 67.5           & 87.3& 88.1  & 93.3 & 94.0              & 89.4 & \textbf{92.9} \\
 2 &   54.3 & 64.4              & 62.5 & 67.9           & 85.5& 88.8 & 92.2 & 94.1              & 88.1 & 91.7 \\
 3 &   43.8 & 65.1              & 60.5 & \textbf{68.0}  & 81.3& \textbf{89.1} & 89.7 & \textbf{94.4}     & 85.3 & 91.4\\
 4 &   24.1 & \textbf{65.4}     & 58.1 & 67.4           & 74.6& \textbf{89.1} & 84.3 & \textbf{94.4}     & 85.0 & 90.3\\
 5 &   15.9 & \textbf{65.4}     & 54.4 & 67.1           & 66.7& 89.0 & 76.7 & 94.3     & 82.3 & 91.3  \\
\hline
\end{tabular}
\caption{\label{tab:scalePyramid}Comparison of different scale levels for MIFS.}
\end{table*}

\begin{table*}
\centering
\footnotesize
\begin{tabular}{|l c |l c |l c|l c|l c|}
\hline
    \multicolumn{2}{|c|}{HMDB51 (MAcc. $\%$)} & \multicolumn{2}{|c|}{Hollywood2 (MAP $\%$)} & \multicolumn{2}{|c|}{UCF101(MAcc. $\%$)} & \multicolumn{2}{|c|}{UCF50 (MAcc. $\%$)} & \multicolumn{2}{|c|}{Olympics Sports (MAP $\%$)}\\ \hline
    
Oneata \textit{et al.} \cite{oneata2013action}  &54.8   &  Lin \textit{et al.} \cite{sun2014dl}  & 48.1 &  Karpathy \textit{et al.} \cite{karpathy2014large}  & 65.4 & Sanath \textit{et al.} \cite{narayan2014cause}  & 89.4 & Jain \textit{et al.} \cite{jain2013better}  & 83.2\\
Wang  \textit{et al.} \cite{wang2013action}   &57.2   & Sapienz \textit{et al.} \cite{sapienza2014feature}  & 59.6   & Sapienz \textit{et al.} \cite{sapienza2014feature}  & 82.3 & Arridhana \textit{et al.} \cite{ciptadi2014movement}  & 90.0  & Adrien \textit{et al.} \cite{gaidon2014activity}  & 85.5 \\
Simonyan \textit{et al.} \cite{simonyan2014two}  & 59.4 &Jain \textit{et al.} \cite{jain2013better}    & 62.5  & Wang \textit{et al.} \cite{wang2013lear}  & 85.9  & Oneata \textit{et al.} \cite{oneata2013action} &90.0 &   Oneata \textit{et al.} \cite{oneata2013action}  & 89.0 \\
Peng \textit{et al.} \cite{peng2014bag}  & 61.1 &  Oneata \textit{et al.} \cite{oneata2013action} &  63.3   &Peng \textit{et al.} \cite{peng2014bag}  &  87.9  & Wang \& Schmid \cite{wang2013action}  & 91.2 &  Wang \& Schmid \cite{wang2013action}  & 91.1\\
Peng \textit{et al.} \cite{peng2014action}  & \textbf{66.8} & Wang \textit{et al.} \cite{wang2013action}  &64.3  &Simonyan \textit{et al.}  \cite{simonyan2014two}   & 88.0  & Peng \textit{et al.} \cite{peng2014bag}  &  92.3  &  Peng \textit{et al.} \cite{peng2014action}   & \textbf{93.8}\\
\hline 
MIFS (L=3)  & 65.1 & MIFS (L = 3) &  \textbf{68.0} & MIFS (L = 3) & \textbf{89.1} & MIFS (L=3)  & \textbf{94.4} & MIFS (L = 3) &  91.4\\ 
\hline
\end{tabular}
\caption{\label{tab:state-of-art}Comparison of our results to the state-of-the-arts.}
\end{table*}

\paragraph{Comparing with State-of-the-Arts}
In Table \ref{tab:state-of-art}, we  compare MIFS at $L=3$, which performs well across all  action datasets, with the state-of-the-art approaches. From Table \ref{tab:state-of-art}, in most of the datasets, we observe improvement over the state-of-the-arts except hmdb51 and Olympics Sports, on which our $L=3$ MIFS give inferior performance . Note that although we list several most recent approaches here for comparison purposes, \textit{most of them are not directly comparable to our results due to the use of different features and representations}. The most comparable one is Wang \& Schmid. \cite{wang2013action}, from which we build our approaches on.  Sapienz \textit{et al.} \cite{sapienza2014feature} explored ways to sub-sample and generate vocabularies for Dense Trajectory features. 
Jain et al. \cite{jain2013better}'s approach incorporated  a new motion descriptor. Oneata et al.  \cite{oneata2013action} focused on testing Spatial Fisher Vector for multiple action and event tasks. Peng et al. \cite{peng2014bag}  improved the performance of Improved Dense Trajectory by increasing the codebook size and fusing multiple coding methods. Karpathy et al. \cite{karpathy2014large} trained a deep convolutional neural network using 1 million weakly labeled YouTube videos and reported  65.4\% mean accuracy on UCF101 datasets. Simonyan \& Zisserman \cite{simonyan2014two} reported results that are competitive to Improved Dense Trajectory by training deep convolutional neural networks using both sampled frames and optical flows and get $57.9\%$ MAcc in HMDB51 and $87.6\%$ MAcc in UCF101, which are comparable to the results of Wang $\&$ Schmid. Peng et al. \cite{peng2014action} achieves better results than us on HMDB51 and Olympic Sports datasets by combining a hierarchical Fisher Vector with the original one.  

\begin{table*}
\centering
\begin{tabular}{| c | c | c| c| c | c | c |}
\hline
& HMDB51 & Hollywood2  & UCF101 & UCF50  & Olympics Sports  & Computational Cost\\ 
& (MAcc$\%$) & (MAP$\%$) & (MAcc$\%$) & (MAcc$\%$) & (MAP$\%$) & (Relative)\\\hline
  L=0 & 62.1 & 67.0 & 87.3 & 93.0 & 89.8 & 1.0\\ 
  L=1-0 & 63.1 & 66.4 & 87.3 & 93.3 & 89.4 & 0.5\\
  L=2-0 & 63.9 & 67.6 & 88.5 & 93.8 & 91.9  & 0.75\\
\hline
\end{tabular}
\caption{\label{tab:time}Performance versus relative computational cost for feature extraction}
\end{table*}

\subsection{Event Detection}
\paragraph{Problem Formulation}
Given a collection of videos, the goal of an event detection task is to detect events of interest such as \textit{Birthday Party} and \textit{Parade}, solely based on the video content. The task is very challenging due to complex actions and scenes. By evaluating on this task, we examine whether MIFS can improve the performance of recognizing very complex actions. 

\paragraph{Dataset}
TREC Video Retrieval Evaluation (TRECVID) Multimedia Event Detection (MED) \cite{over2013trecvid} is a task organized by NIST (National Institute of Standards and Technology) aimed at encouraging new technologies for detecting complex events such as \textit{having a birthday party}. Started in 2010, NIST has gradually built up a database that contains 8000 hours of videos and 40 events, which is by far the largest event detection collection. MEDTEST13, 14 datasets are two standard system evaluation datasets released by NIST in 2013 and 2014, respectively. Each of them contains around 10 percent of the whole MED collection and has 20 events. They consist of two tasks, i.e. EK100 and EK10. EK100 task has 100 positive training samples while EK10 has 10. For both tasks, they have around 5000 background samples. Together, each dataset has 8000 training samples and 24000 testing samples. 

\paragraph{Experimental Setting}A similar setting discussed in section \ref{action_recognition} is applied except we use five folders cross-validation to choose the penalty parameter C for linear SVM.  For each classifier, C is chosen among $10^{-3}, 10^{-2},10^{-1},1,10^{1},10^{2},10^{3}$.  We only test MIFS with $L=3$ as recommended in section \ref{action_recognition} because extracting Dense Trajectory feature from such large datasets itself is very time consuming. It took us 4 days to generate representations for both MEDTEST13, 14 using a cluster with more than 500 Intel E565+ series processors. We use  MAP as evaluation criteria.

\paragraph{Results} Table \ref{tab:med} lists the overall MAP (detail results can be found in supplementary materials). The baseline method is a conventional single-scale representation with $L=0$. From Table \ref{tab:med}, we can see that for both MEDTEST13 and MEDTEST14, MIFS representations consistently improve over the original-scale representation by about $2\%$ in both EK100 and EK10. It is worth emphasizing that MED is such a challenging task that $2\%$ of absolute performance improvement is quite significant.

\begin{table}
\centering
\begin{tabular}{|c | c c|c c|}
\hline
 & \multicolumn{2}{|c|}{MEDTEST13}  & \multicolumn{2}{|c|}{MEDTEST14}\\
\hline 
 & EK100& EK10  & EK100 & EK10 \\\hline
  Baseline & 34.2 & 17.7& 27.3  & 12.7  \\ 
  MIFS (L=3) & \textbf{36.3}& \textbf{19.3} &  \textbf{29.0} & \textbf{14.9}  \\
\hline
\end{tabular}
\caption{\label{tab:med}Performance Comparison on the MED task.}
\end{table}

\subsection{Computational Complexity}
Level 0 of a MIFS representation has the same cost as other single pass methods, e.g., Wang \& Schmid. \cite{wang2013action}.  For level $l$, the cost becomes $1/l$ of the level 0. So with a MIFS up to level 2, the computational cost will be less than twice the cost of a single pass through the original video, yet it can significantly improve the single-pass methods. If computational efficiency is critical, the method can be sped up by removing low-scale features. For example, removing L=0 (original videos) will significantly reduce cost but still give useful improvements as shown in Table \ref{tab:time}. $L-1$ shows the results of only using features from every 2nd frame and $L=2-0$ shows the results of combining features from level 1 (every 2nd frame) and level 2 (every 3rd frame) but not L=0. As seen, in most of cases, we can still get better results with less cost.

\section{Conclusion}

We develop the Multi-skIp Feature Stacking (MIFS) method for enhancing the learnability of action representations. MIFS stacks features extracted using a family of differential filters parameterized with multiple time skips and achieves shift-invariance in the frequency space. In contrast to Gaussian Pyramid, MIFS generates features at all scales and tends to cover a longer range of action signals. Theoretical results show that MIFS improves the learnability of action representation exponentially. Extensive experiments on seven real-world datasets show that MIFS exceeds state-of-the-art methods. Future works would be determining the appropriate level for different action types. Additionally, we would like to improve the quality of optical flow calculation and tracking at coarse scales. 

\section{Acknowledgement}
This work was partially supported by Intelligence
Advanced Research Projects Activity (IARPA)
via Department of Interior National Business Center contract
number D11PC20068. The U.S. Government is authorized
to reproduce and distribute reprints for Governmental
purposes notwithstanding any copyright annotation
thereon. Disclaimer: The views and conclusions contained
herein are those of the authors and should not be interpreted
as necessarily representing the official policies or endorsements,
either expressed or implied, of IARPA, DoI/NBC, or
the U.S. Government.
The work was also supported in part by the U. S. Army Research Office (W911NF-13-1-0277). Any opinions, findings, and conclusions or recommendations expressed in this material are those of the authors and do not necessarily reflect the views of ARO.
{\small
\bibliographystyle{ieee}
\bibliography{egbib}

\begin{thebibliography}{10}\itemsep=-1pt

\bibitem{adelson1984pyramid}
E.~H. Adelson, C.~H. Anderson, J.~R. Bergen, P.~J. Burt, and J.~M. Ogden.
\newblock Pyramid methods in image processing.
\newblock {\em RCA engineer}, 29(6):33--41, 1984.

\bibitem{aggarwal2011human}
J.~Aggarwal and M.~S. Ryoo.
\newblock Human activity analysis: A review.
\newblock {\em ACM Computing Surveys (CSUR)}, 43(3):16, 2011.

\bibitem{chu2005foundations}
W.~Chu and T.~Y. Lin.
\newblock {\em Foundations and advances in data mining}, volume 180.
\newblock Springer, 2005.

\bibitem{ciptadi2014movement}
A.~Ciptadi, M.~S. Goodwin, and J.~M. Rehg.
\newblock Movement pattern histogram for action recognition and retrieval.
\newblock In {\em ECCV}. 2014.

\bibitem{darrell1993space}
T.~Darrell and A.~Pentland.
\newblock Space-time gestures.
\newblock In {\em CVPR}, 1993.

\bibitem{gaidon2014activity}
A.~Gaidon, Z.~Harchaoui, and C.~Schmid.
\newblock Activity representation with motion hierarchies.
\newblock {\em International Journal of Computer Vision}, 107(3):219--238,
  2014.

\bibitem{jain2013better}
M.~Jain, H.~J{\'e}gou, and P.~Bouthemy.
\newblock Better exploiting motion for better action recognition.
\newblock In {\em CVPR}, 2013.

\bibitem{jiang2012trajectory}
Y.-G. Jiang, Q.~Dai, X.~Xue, W.~Liu, and C.-W. Ngo.
\newblock Trajectory-based modeling of human actions with motion reference
  points.
\newblock In {\em ECCV}. 2012.

\bibitem{karpathy2014large}
A.~Karpathy, G.~Toderici, S.~Shetty, T.~Leung, R.~Sukthankar, and L.~Fei-Fei.
\newblock Large-scale video classification with convolutional neural networks.
\newblock In {\em CVPR}, 2014.

\bibitem{koenderink1984structure}
J.~J. Koenderink.
\newblock The structure of images.
\newblock {\em Biological cybernetics}, 50(5):363--370, 1984.

\bibitem{kuehne2011hmdb}
H.~Kuehne, H.~Jhuang, E.~Garrote, T.~Poggio, and T.~Serre.
\newblock Hmdb: a large video database for human motion recognition.
\newblock In {\em ICCV}, 2011.

\bibitem{laptev2005space}
I.~Laptev.
\newblock On space-time interest points.
\newblock {\em IJCV}, 64(2-3):107--123, 2005.

\bibitem{lindeberg1993detecting}
T.~Lindeberg.
\newblock Detecting salient blob-like image structures and their scales with a
  scale-space primal sketch: a method for focus-of-attention.
\newblock {\em IJCV}, 11(3):283--318, 1993.

\bibitem{lindeberg1994linear}
T.~Lindeberg and B.~M. ter Haar~Romeny.
\newblock {\em Linear scale-space I: Basic theory}.
\newblock Springer, 1994.

\bibitem{lowe2004distinctive}
D.~G. Lowe.
\newblock Distinctive image features from scale-invariant keypoints.
\newblock {\em IJCV}, 60(2):91--110, 2004.

\bibitem{marr1982vision}
D.~Marr.
\newblock Vision: A computational investigation into the human representation
  and processing of visual information, henry holt and co.
\newblock {\em Inc., New York, NY}, pages 2--46, 1982.

\bibitem{marszalek2009actions}
M.~Marszalek, I.~Laptev, and C.~Schmid.
\newblock Actions in context.
\newblock In {\em CVPR}, 2009.

\bibitem{matikainen2009trajectons}
P.~Matikainen, M.~Hebert, and R.~Sukthankar.
\newblock Trajectons: Action recognition through the motion analysis of tracked
  features.
\newblock In {\em ICCV Workshops}, 2009.

\bibitem{narayan2014cause}
S.~Narayan and K.~R. Ramakrishnan.
\newblock A cause and effect analysis of motion trajectories for modeling
  actions.
\newblock In {\em CVPR}, 2014.

\bibitem{niebles2010modeling}
J.~C. Niebles, C.-W. Chen, and L.~Fei-Fei.
\newblock Modeling temporal structure of decomposable motion segments for
  activity classification.
\newblock In {\em ECCV}. 2010.

\bibitem{oneata2013action}
D.~Oneata, J.~Verbeek, C.~Schmid, et~al.
\newblock Action and event recognition with fisher vectors on a compact feature
  set.
\newblock In {\em ICCV}, 2013.

\bibitem{over2013trecvid}
P.~Over, G.~Awad, J.~Fiscus, and G.~Sanders.
\newblock Trecvid 2013--an introduction to the goals, tasks, data, evaluation
  mechanisms, and metrics.
\newblock 2013.

\bibitem{park2004hierarchical}
S.~Park and J.~K. Aggarwal.
\newblock A hierarchical bayesian network for event recognition of human
  actions and interactions.
\newblock {\em Multimedia systems}, 10(2):164--179, 2004.

\bibitem{peng2014bag}
X.~Peng, L.~Wang, X.~Wang, and Y.~Qiao.
\newblock Bag of visual words and fusion methods for action recognition:
  Comprehensive study and good practice.
\newblock {\em arXiv preprint arXiv:1405.4506}, 2014.

\bibitem{peng2014action}
X.~Peng, C.~Zou, Y.~Qiao, and Q.~Peng.
\newblock Action recognition with stacked fisher vectors.
\newblock In {\em Computer Vision--ECCV 2014}, pages 581--595. Springer, 2014.

\bibitem{perronnin2010improving}
F.~Perronnin, J.~S{\'a}nchez, and T.~Mensink.
\newblock Improving the fisher kernel for large-scale image classification.
\newblock In {\em ECCV}. 2010.

\bibitem{reddy2013recognizing}
K.~K. Reddy and M.~Shah.
\newblock Recognizing 50 human action categories of web videos.
\newblock {\em Machine Vision and Applications}, 24(5):971--981, 2013.

\bibitem{sapienza2014feature}
M.~Sapienza, F.~Cuzzolin, and P.~H. Torr.
\newblock Feature sampling and partitioning for visual vocabulary generation on
  large action classification datasets.
\newblock {\em arXiv preprint arXiv:1405.7545}, 2014.

\bibitem{sermanet2013overfeat}
P.~Sermanet, D.~Eigen, X.~Zhang, M.~Mathieu, R.~Fergus, and Y.~LeCun.
\newblock Overfeat: Integrated recognition, localization and detection using
  convolutional networks.
\newblock {\em arXiv preprint arXiv:1312.6229}, 2013.

\bibitem{zhen2013spatio}
L.~Shao, X.~Zhen, D.~Tao, and X.~Li.
\newblock Spatio-temporal laplacian pyramid coding for action recognition.
\newblock {\em IEEE Transactions on Cybernetics}, pages 2168--2267, 2013.

\bibitem{simonyan2014two}
K.~Simonyan and A.~Zisserman.
\newblock Two-stream convolutional networks for action recognition in videos.
\newblock {\em arXiv preprint arXiv:1406.2199}, 2014.

\bibitem{simonyan2014very}
K.~Simonyan and A.~Zisserman.
\newblock Very deep convolutional networks for large-scale image recognition.
\newblock {\em arXiv preprint arXiv:1409.1556}, 2014.

\bibitem{soomro2012ucf101}
K.~Soomro, A.~R. Zamir, and M.~Shah.
\newblock Ucf101: A dataset of 101 human actions classes from videos in the
  wild.
\newblock {\em arXiv preprint arXiv:1212.0402}, 2012.

\bibitem{sun2009hierarchical}
J.~Sun, X.~Wu, S.~Yan, L.-F. Cheong, T.-S. Chua, and J.~Li.
\newblock Hierarchical spatio-temporal context modeling for action recognition.
\newblock In {\em CVPR}, 2009.

\bibitem{sun2014dl}
L.~Sun, K.~Jia, T.-H. Chan, Y.~Fang, G.~Wang, and S.~Yan.
\newblock Dl-sfa: Deeply-learned slow feature analysis for action recognition.
\newblock In {\em CVPR}, 2014.

\bibitem{tropp2012user}
J.~A. Tropp.
\newblock User-friendly tail bounds for sums of random matrices.
\newblock {\em Foundations of Computational Mathematics}, 12(4):389--434, 2012.

\bibitem{wang2011action}
H.~Wang, A.~Klaser, C.~Schmid, and C.-L. Liu.
\newblock Action recognition by dense trajectories.
\newblock In {\em CVPR}, 2011.

\bibitem{wang2013lear}
H.~Wang and C.~Schmid.
\newblock Lear-inria submission for the thumos workshop.
\newblock In {\em ICCV Workshop}, 2013.

\bibitem{wang2013action}
H.~Wang, C.~Schmid, et~al.
\newblock Action recognition with improved trajectories.
\newblock In {\em ICCV}, 2013.

\bibitem{yamato1992recognizing}
J.~Yamato, J.~Ohya, and K.~Ishii.
\newblock Recognizing human action in time-sequential images using hidden
  markov model.
\newblock In {\em CVPR}, 1992.

\end{thebibliography}
}

\onecolumn

\section{Proof}

Here we give the details of proofs in the main text. Our proofs are
based on the following  Bernstein's Matrix Inequaltiy.
\begin{lem}[Bernstein's Matrix Inequality]
\label{lem:Bernstein-ineq} Let $\mathbf{x}_{i}\in\mathbb{R}^{p\times1}$,
$\left\Vert \mathbf{x}_{i}\right\Vert ^{2}\leq B$. $S=\mathbf{x}_{1}\mathbf{x}_{1}{}^{\mathrm{T}}+\cdots\mathbf{x}_{n}\mathbf{x}_{n}{}^{\mathrm{T}}$.
Then with probability at least $1-\delta$, 
\begin{equation}
\left\Vert S-\mathrm{E}\{S\}\right\Vert \leq\sqrt{2B\left\Vert \mathrm{E}\{S\}\right\Vert \log(2p/\delta)}+\frac{B}{3}\log(2p/\delta)\ .
\end{equation}

\end{lem}

\subsection{Proof of Theorem \ref{thm:condition-number-P-under-fixed-tao}}
\begin{proof}
For the $i$-th row, $j$-th column of $P$, 
\begin{align}
|P_{i,j}|= & [\boldsymbol{\alpha}_{i}(t_{j}+\tau)-\boldsymbol{\alpha}_{i}(t_{j})]\leq2\\
\left\Vert P_{j}\right\Vert ^{2}\leq & 4k\\
\mathrm{E}\{P_{i,j}^{2}\}\leq & 2(1+c)\exp(-\gamma/\tau)\\
\mathrm{E}\{P_{i,j}^{2}\}\geq & 2\exp(-\gamma/\tau)\\
\mathrm{E}\{P_{i,j}P_{k,j}\}= & 0\quad i\not=k\\
\lambda_{\max}\{\mathrm{E}\{P_{j}P_{j}{}^{\mathrm{T}}\}\}= & \frac{1}{T}\lambda_{\max}\{\mathrm{E}\{PP{}^{\mathrm{T}}\}\}\leq2(1+c)\exp(-\gamma_{1}/\tau) \label{eq:21}\\
\lambda_{\min}\{\mathrm{E}\{P_{j}P_{j}{}^{\mathrm{T}}\}\}= & \frac{1}{T}\lambda_{\min}\{\mathrm{E}\{PP{}^{\mathrm{T}}\}\}\geq2\exp(-\gamma_{k}/\tau)\label{eq:22}\ .
\end{align}
The equalities in Eq. (\ref{eq:21}) and  Eq. (\ref{eq:22}) come from the fact that $P_j$ is assumed to be an
independent and indentical sample from the column distribution of $P$.

 By  Bernstein's Matrix inequality (Theorem \ref{thm:condition-number-P-under-fixed-tao}),
with probability at least $1-\delta$, we have
\begin{align}
4T\Delta_{\tau}\triangleq\left\Vert PP{}^{\mathrm{T}}-\mathrm{E}\{PP{}^{\mathrm{T}}\}\right\Vert \leq & \sqrt{2\times4k\times2T(1+c)\exp(-\gamma_{1}/\tau)\log(2k/\delta)}+\frac{4k}{3}\log(2k/\delta)\nonumber \\
= & 4\sqrt{kT(1+c)\exp(-\gamma_{1}/\tau)\log(2k/\delta)}+\frac{4}{3}k\log(2k/\delta)\nonumber \\
\leq & 4\sqrt{kT(1+c)\log(2k/\delta)}+\frac{4}{3}k\log(2k/\delta)
\end{align}
 When 
\begin{equation}
T\geq\frac{1}{9(1+c)}k\log(2k/\delta)
\end{equation}
 we have
\begin{equation}
\Delta_{\tau}\leq2\sqrt{k\frac{1}{T}(1+c)\log(2k/\delta)}
\end{equation}

Therefore, when $T$ large enough,
\begin{equation}
\beta(PP{}^{\mathrm{T}})\leq\frac{(1+c)\exp(-\gamma_{1}/\tau)+\Delta_{\tau}}{\exp(-\gamma_{k}/\tau)-\Delta_{\tau}}~.
\end{equation}

A lower bound on $\beta(PP{}^{\mathrm{T}})$ could be given similarly
by changing $\Delta_{\tau}$ to $-\Delta_{\tau}$ .
\end{proof}

\subsection{Proof of Theorem \ref{thm:condition-number-MSP}}
\begin{proof}
The proof is similar to Theorem \ref{thm:condition-number-P-under-fixed-tao},
except that $P_{i}$ is sampled from $m$ different distribution.
To borrow the proof in Theorem \ref{thm:condition-number-P-under-fixed-tao},
the distribution of $P_{i}$ has $m$ components. The $i$-th component
is sampled from $i\tau$ skip with probability $T_{i}/\sum_{j}T_{j}=T_{i}/T$
where $T=\sum_{j}T_{j}$ is the total number of features. Based on
this observation, we have:
\begin{align}
|P_{i,j}|\leq & 2\\
\left\Vert P_{j}\right\Vert ^{2}\leq & 4k\\
\mathrm{E}\{P_{i,j}^{2}\}\leq & \sum_{i}\frac{T_{i}}{T}2(1+c)\exp(-\gamma/\tau_{i})\\
\mathrm{E}\{P_{i,j}^{2}\}\geq & \sum_{i}\frac{T_{i}}{T}2\exp(-\gamma/\tau_{i})\\
\mathrm{E}\{P_{i,j}P_{k,j}\}= & 0\quad i\not=k\\
\lambda_{\max}\{\mathrm{E}\{P_{j}P_{j}{}^{\mathrm{T}}\}\}= & \frac{1}{T}\lambda_{\max}\{\mathrm{E}\{PP{}^{\mathrm{T}}\}\}\leq\sum_{i}\frac{T_{i}}{T}2(1+c)\exp(-\gamma_{1}/\tau_{i})\\
\lambda_{\min}\{\mathrm{E}\{P_{j}P_{j}{}^{\mathrm{T}}\}\}= & \frac{1}{T}\lambda_{\min}\{\mathrm{E}\{PP{}^{\mathrm{T}}\}\}\geq\sum_{i}\frac{T_{i}}{T}2\exp(-\gamma_{k}/\tau_{i})
\end{align}
 From matrix concentration,

\begin{align}
4T\Delta_{\tau}\triangleq\left\Vert PP{}^{\mathrm{T}}-\mathrm{E}\{PP{}^{\mathrm{T}}\}\right\Vert \leq & \sqrt{2\times4k\times[\sum_{i}T_{i}2(1+c)\exp(-\gamma/\tau_{i})]\log(2k/\delta)}+\frac{4k}{3}\log(2k/\delta)\nonumber \\
= & 4\sqrt{k[\sum_{i}T_{i}(1+c)\exp(-\gamma/\tau_{i})]\log(2k/\delta)}+\frac{4}{3}k\log(2k/\delta)\nonumber \\
\leq & 4\sqrt{k(1+c)T\log(2k/\delta)}+\frac{4}{3}k\log(2k/\delta)
\end{align}
 When 
\begin{equation}
T\geq\frac{1}{9(1+c)}k\log(2k/\delta)
\end{equation}
 we have
\begin{align}
\Delta_{\tau}\leq & 2\sqrt{k\frac{1}{T}(1+c)\log(2k/\delta)}\\
\beta(PP{}^{\mathrm{T}})\leq & \frac{\sum_{i}\frac{T_{i}}{T}2(1+c)\exp(-\gamma_{1}/\tau_{i})+\Delta_{\tau}}{\sum_{i}\frac{T_{i}}{T}2\exp(-\gamma_{k}/\tau_{i})-\Delta_{\tau}}~.
\end{align}
\end{proof}

\end{document}